\pdfoutput=1

\documentclass[11pt]{article}

\usepackage[preprint]{acl}

\usepackage{times}
\usepackage{latexsym}
\usepackage{booktabs}
\usepackage{amsmath}
\usepackage{tabularx}
\usepackage{arydshln}
\usepackage{multirow} 
\usepackage{bbding}
\usepackage{pifont}
\usepackage{listings}

\usepackage[T1]{fontenc}

\usepackage[utf8]{inputenc}

\usepackage{microtype}

\usepackage{inconsolata}

\usepackage{graphicx}

\title{Make Some Noise: Unlocking Language Model Parallel Inference Capability through Noisy Training}

\author{
 \textbf{Yixuan Wang\textsuperscript{1,}\footnotemark[2]},
 \textbf{Xianzhen Luo\textsuperscript{1,}\footnotemark[2]},
 \textbf{Fuxuan Wei\textsuperscript{1}},
 \textbf{Yijun Liu\textsuperscript{1}},
 \textbf{Qingfu Zhu\textsuperscript{1,}\footnotemark[1]},
\\
 \textbf{Xuanyu Zhang\textsuperscript{2}},
 \textbf{Qing Yang\textsuperscript{2}},
 \textbf{Dongliang Xu\textsuperscript{2}},
 \textbf{Wanxiang Che\textsuperscript{1}}
\\
 \textbf{\textsuperscript{1}} Harbin Institute of Technology, Harbin, China 
 \\
 \textbf{\textsuperscript{2}} Du Xiaoman (Beijing) Science Technology Co., Ltd.
 \\
 \texttt{\{wyx,xzluo,fxwei,yjliu,qfzhu,car\}@ir.hit.edu.cn}
 \\
 \texttt{\{zhangxuanyu,yangqing,xudongliang\}@duxiaoman.com}
}

\begin{document}
\maketitle

\renewcommand{\thefootnote}{\fnsymbol{footnote}}
\footnotetext[2]{Equal contribution.}
\footnotetext[1]{Corresponding author.}
\renewcommand{\thefootnote}{\arabic{footnote}}

\begin{abstract}

Existing speculative decoding methods typically require additional model structure and training processes to assist the model for draft token generation.
This makes the migration of acceleration methods to the new model more costly and more demanding on device memory.
To address this problem, we propose the \textbf{Make Some Noise} (MSN) training framework as a replacement for the supervised fine-tuning stage of the large language model.
The training method simply introduces some noise at the input for the model to learn the denoising task.
It significantly enhances the parallel decoding capability of the model without affecting the original task capability.
In addition, we propose a tree-based retrieval-augmented Jacobi (TR-Jacobi) decoding strategy to further improve the inference speed of MSN models.
Experiments in both the general and code domains have shown that MSN can improve inference speed by 2.3-2.7x times without compromising model performance.
The MSN model also achieves comparable acceleration ratios to the SOTA model with additional model structure on Spec-Bench.
\end{abstract}

\section{Introduction}

Large language models (LLMs) represented by GPT-4 \cite{openai2024gpt4} and LLaMA \cite{touvron2023llama} have made great breakthroughs to artificial intelligence \cite{kocon2023chatgpt}.
However, LLMs suffer from high inference latency due to the autoregressive (AR) decoding paradigm, which constrains the model to generate only one token per decoding step.
It significantly limits the applications of LLMs when needs lengthy response.

To address the bottleneck introduced by AR, 
speculative decoding \cite{leviathan2023fast,chen2023accelerating} is proposed to get more than one token in one decoding step.
It first guesses multi-step draft tokens and then verifies them simultaneously in one model forward.
Once any draft token is accepted, it can effectively speedup the inference process.
\citet{chen2023accelerating} employ a relatively small LLM to generate multi-step draft tokens and verify them in parallel on the target LLM.
Medusa \cite{cai2024medusa} extends and train multiple language model heads for existing models to predict later draft tokes.
It achieves considerable inference speedup through efficient validation using tree attentions.
BiTA \cite{lin2024bita} takes full advantage of the capabilities of LLM itself through a parameter-efficient design that allows the model to generate daft tokens based on trainable special tokens.
\citet{kou2024cllms} propose a post-training method based on constructed Jacobi trajectories that can accelerate the model's own Jacobi decoding capabilities.

\begin{figure}[t]
    \centering
    \includegraphics[width=\columnwidth]{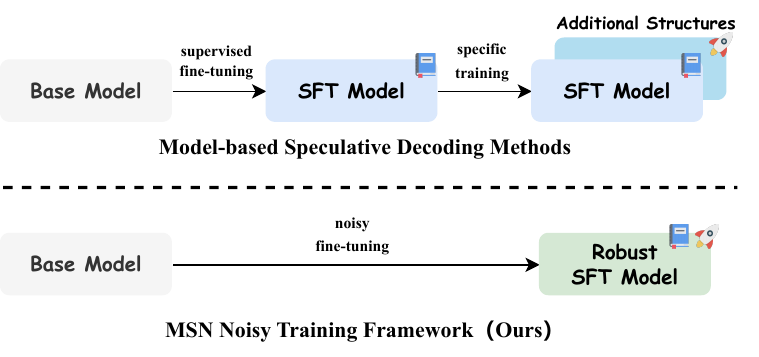}
    \caption{An illustration of the differences between the proposed MSN framework and existing model-based speculative decoding methods.
    The book icon represents task-specific capabilities and the rocket icon represents parallel decoding capabilities.}
    \label{fig:intro}
\end{figure}

Although the above methods improve the inference efficiency of the model to a certain extent, there are still some problems to be solved as shown in Figure \ref{fig:intro}.
(1) \textbf{Additional Structures}.
Most current speculative decoding methods rely heavily on additional model structures to accomplish draft token prediction (e.g., separate models, language model heads, trainable prompts, etc.).
In the case of Medusa, for example, it adds 1.6B parameters (5 additional medusa heads) to the 7B target model, which will undoubtedly increase the memory requirements for model inference.
(2) \textbf{Separate Post-Training}.
Existing model-based speculative decoding methods are trained after LLMs' supervised fine-tuning (SFT) stage to obtain acceleration capability.
This process usually requires complex model setups or time-consuming data construction, and some methods even lose part of the model's original task capabilities.
Separate training of task and acceleration capabilities leads to an overly complex approach which is not easy to deploy.

To address the above problem, we propose a noisy training framework \footnote{\href{https://github.com/wyxstriker/MakeSomeNoiseInference}{https://github.com/wyxstriker/MakeSomeNoiseInference}} \textbf{M}ake \textbf{S}ome \textbf{N}oise (\textbf{MSN}) as a replacement for SFT, which enables the model to acquire both task-relevant capability as well as acceleration capability at the same stage without the need for additional structures and training stages.
Specifically, we consider the process of Jacobi decoding \cite{santilli2023accelerating} as a denoising process, and improve the denoising ability of the model by including a causal language model denoising task in the SFT stage.
Since the SFT stage is almost a necessary aspect of LLM applications, our proposed approach can be interpreted as a \textbf{free lunch} to the parallel inference capability of LLMs.
In the inference phase, we use Jacobi decoding to achieve inference acceleration through repeated iterations of random noise tokens as well as verification.
Besides, in order to alleviate the cold-start problem of Jacobi decoding and mitigate the effect of random initial noise, we also propose the tree-based retrieval-augmented Jacobi 
(TR-Jacobi) decoding method, which can effectively improve the speedup ratio.

We have conducted detailed experiments in the general and code domains. The results show that the MSN training framework can significantly improve the denoising ability of the model without affecting the performance of the original SFT model, which in turn achieves a 2.3-2.7x inference acceleration effect.
In addition, we performed a detailed evaluation on Specbench, which is specifically designed for speculative decoding.
As a speculative decoding method without additional structure and training, the acceleration ratio of the MSN model under TR-Jacobi decoding strategy significantly outperforms other additional-structure-free methods and possesses comparable speedup ratios to the SOTA model with additional model structure and training.

Our main contributions can be summarised as follows:
\begin{itemize}
\item We propose a new training framework Make Some Noise (MSN) as an alternative to SFT, which can unlock the parallel decoding capability of the model through the denoising task.
\item We propose a tree-based retrieval-augmented decoding method that effectively improves the inference speed of MSN models under memory bottlenecks.
\item  Experiments show that MSN training enables the model to have a comparable acceleration ratio to the SOTA method without significant loss of task performance.
\end{itemize}

\section{Related Work}
\subsection{Jacobi Decoding}
Jacobi decoding \cite{santilli2023accelerating} treats greedy decoding of generative tasks as solving equations:
\begin{equation}
\begin{aligned}
\begin{cases}
y_{1} &= \arg\max P_{\theta}(y_1|x) \\
y_{2} &= \arg\max P_{\theta}(y_2|y_1, x) \\
& \vdots \\
y_{m} &= \arg\max P_{\theta}(y_m|y_{1:m-1}, x) \\
\end{cases}
\end{aligned}
\label{eq:jacobi}
\end{equation}
Auto-regressive decoding solves the equations from first to last based on the given input $x$, progressively replacing the resolved variables.
In contrast, Jacobi decoding relies on Jacobi and Gauss-Seidel (GS) fixed-point iteration methods \cite{ortega2000iterative} to solve Equation \ref{eq:jacobi} in parallel.
Specifically, it passes an initialisation sequence of length $m$ into the model for iterative generation until the sequence converges to a fixed point.
Jacobi decoding expects to solve the equation in less than $m$ iterations, but in fact existing models perform poorly under this decoding strategy due to the lack of denoising capability.
\citet{kou2024cllms} greatly improve the efficiency of Jacobi decoding by constructing the trajectory data during Jacobi decoding and performing consistency training.

\begin{figure*}[ht]
  \centering
  \includegraphics[width=0.8\linewidth]{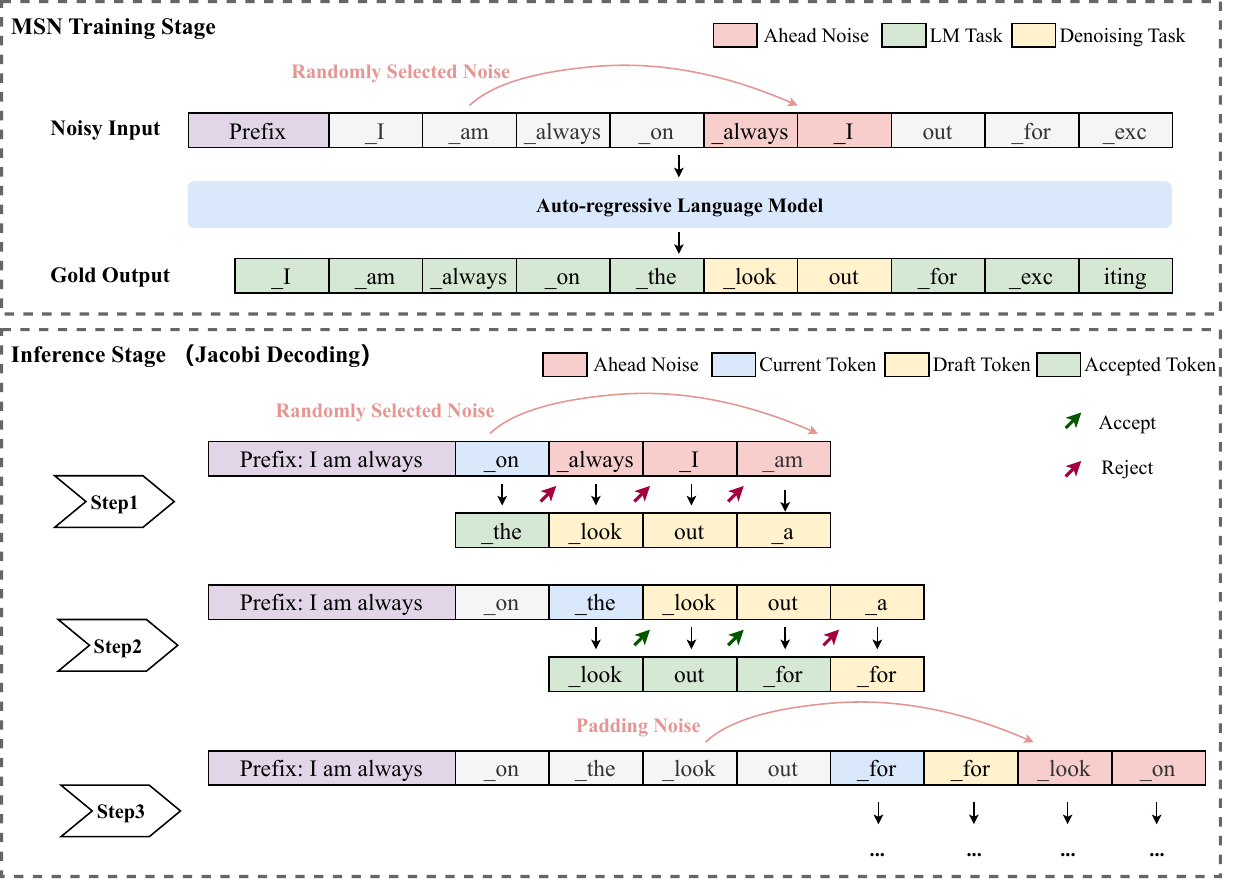}
  \caption {Illustrations of the Make Some Noisy training framework and Jacobi decoding strategy.
  The training phase in the figure uses a noise segment of length 2, and the inference phase is shown as an example when the length of the noise segment is set to 3.
  }
  \label{fig:overall}
\end{figure*}

\subsection{Speculative Decoding}
\label{tree-verify}
Speculative decoding can effectively increase the decoding speed without changing the output quality by guessing and verifying the output of the auto-regressive language model in parallel.
Current mainstream work has focused on investigating how to complete draft token generation efficiently.
\citet{stern2018blockwise} complete the prediction of draft tokens with additional model structures.
\citet{chen2023accelerating} generate reliable draft tokens by a external small model.
\citet{cai2024medusa} train multiple heads for the LLM model for predicting draft tokens based on the previous work.
\citet{li2024eagle} make full use of the information in the hidden layer to accomplish high-quality predictions of draft models with a separate decoder layer.
\citet{lin2024bita} enable the model to predict draft tokens by training prefix tokens.

In addition, there are some speculative decoding methods that do not require training.
LLMA \cite{yang2023inference} achieves 2x$\sim$3x speedups on tasks such as conversations by retrieving text segments from reference texts.
\citet{fu2024break} performs more efficient verification by collecting n-gram segments generated during Jacobi decoding as draft tokens.
\citet{saxena2023prompt} achieves acceleration in specific domains simply by retrieving draft tokens from the ahead prompt.
REST \cite{he2023rest} enables plug-in draft token generation by retrieving a constructed knowledge database.
\citet{zhao2024ouroboros} proposes Ouroboros that combines the advantages of both the retrieval and draft model approaches.
It utilizes the retrieval method to further enhance the generation length of the draft model, achieving a significant speedup ratio.

In order to further improve the verification efficiency of draft token, \citet{miao2023specinfer} propose to verify multiple paths as a token tree at a time by designing an attention mask matrix.
Nowadays, token tree verification has become a widely used technique to improve the verification efficiency of speculative decoding.

\section{Method}

\subsection{Overall}
\label{method_overall}
Our core idea is to consider parallel decoding as a kind of text generation under noise, similar to the Jacobi decoding.
This requires the model to have the ability to generate the corresponding correct token despite the noisy token, which is not possible with the current teacher-forcing training \cite{bachmann2024pitfalls}.

Inspired by related work addressing exposure bias \cite{bengio2015scheduled,zhang2019bridging}, we chose to enhance the denoising ability of the model by adding some token-level noise to the input sequence in the SFT stage of LLMs.
As shown in Figure \ref{fig:overall}, we incorporate a causal language model denoising task in the training phase to ensure that the model has the robust generation capability.
During the inference phase, we use random noise spliced at the end of the sequence, and keep generating and verifying draft tokens by iterative denoising, consistent with the Jacobi decoding process.

To guarantee that the denoising ability is improved without affecting the acquisition of task capabilities, we construct the method in terms of the content and location of the noise segment(Section \ref{noisy_training}).
In addition, to further enhance the validation efficiency of the model, we propose a tree-based retrieval-augmented Jacobi (TR-Jacobi) decoding strategy (Section \ref{infernce}).

\subsection{Noisy Training Framework}
\label{noisy_training}
Teacher-forcing has been widely adopted as an efficient training method by the dominant generative models.
It trains the model with the label at moment $t$ as the input at moment $t+1$, which can accelerate the model convergence.
For the sequence $X={x_0x_1...x_n}$, the loss function of a traditional auto-regressive model can be formulated as:
\begin{equation}
  \label{eq:ar_loss}
  Loss_{AR} = \sum_{i=0}^{n} -\log P(X_i|X_{<i};\theta)
\end{equation}
where $\theta$ is the set of parameters of the language model and $X_{<i}$ represents the sub-sequence $x_0x_1...x_{i-1}$.
The model is trained to generate results based on the correct labels, therefore each generation step requires the results generated in the previous step.

In order to equip the model with denoising capability, we introduce causal noise token in the training phase.
As shown in Figure \ref{fig:overall}, we insert some noise tokens at the input to break the restriction that teacher-forcing always takes golden labels as input.
To minimise the impact of noise on training, we only replace one short segment with noise tokens in each sample.
The noise sample can be expressed as
$\hat{X}={x_0}{x_1}...{\hat{x}_i}...{\hat{x}_{j}}...{x_n}$
, where $\hat{x}_i...\hat{x}_{j}$ represents the noise segment.
The loss function of the noisy training method can be formulated as:
\begin{equation}
  \label{eq:msn_loss}
  Loss_{MSN} = \sum_{i=0}^{n} -\log P(X_i|\hat{X}_{<i};\theta)
\end{equation}
where $X_i$ represents the token of golden labels and $\hat{X}_{<i}$ represents the sub-sequence with noise tokens.
It should be noted that even though the input contains partially noise tokens, the target of the model to learn is still the correct labels.
Such training with noise can unlock the parallel decoding capability of the model to some extent.
To further reduce the impact of noise on the SFT task, we investigate the content of the noise and the location of the noise.
\paragraph{The Content of the Noise Segment.}
The main motivation for noisy training is to equip the model with the ability to generate correct tokens despite noisy inputs, which is achieved through the loss of the noise segments.
However, the causal attention mask of the LLMs leads to the possibility that the noise tokens may have an impact on the later auto-regressive training objectives.
To minimise the impact, we chose the ahead noise as the main content of the segments.
Specifically, we randomly sample the ahead tokens as the current noise token, which can be formulated as:
\begin{equation}
  \label{eq:ahead_noise}
  \hat{x}_i = random\_sample(X_{<i})
\end{equation}
where $X_{<i}$ represents for the sub-sequence ahead of $x_i$.
Compared to random noise, ahead noise has less impact on subsequent tokens.
In addition, denoising the ahead noise tokens is more challenging since they are more relevant to the context.


\paragraph{The Location of Noise Segment.}
Inspired by \citet{lin2024rho}, we have tried two noise location selection methods, random selection and PPL-based selection.
%
Experiments (see the Appendix \ref{app:location} for details) have found that neither method has a significant impact on the model task performance and the speedup ratios are similar.
We speculate that our noise segments (less than 10) may be relatively short on SFT datasets with an average length of 600 or more, and do not have an impact on the training of the model itself.
We therefore choose the simpler random replacement noise method.

In practice, at each step of training, we only replace one fixed-length random segment with ahead noise for the response of each sample.

\subsection{TR-Jacobi Decoding}
\label{infernce}
\begin{figure}[t]
  \includegraphics[width=\linewidth]{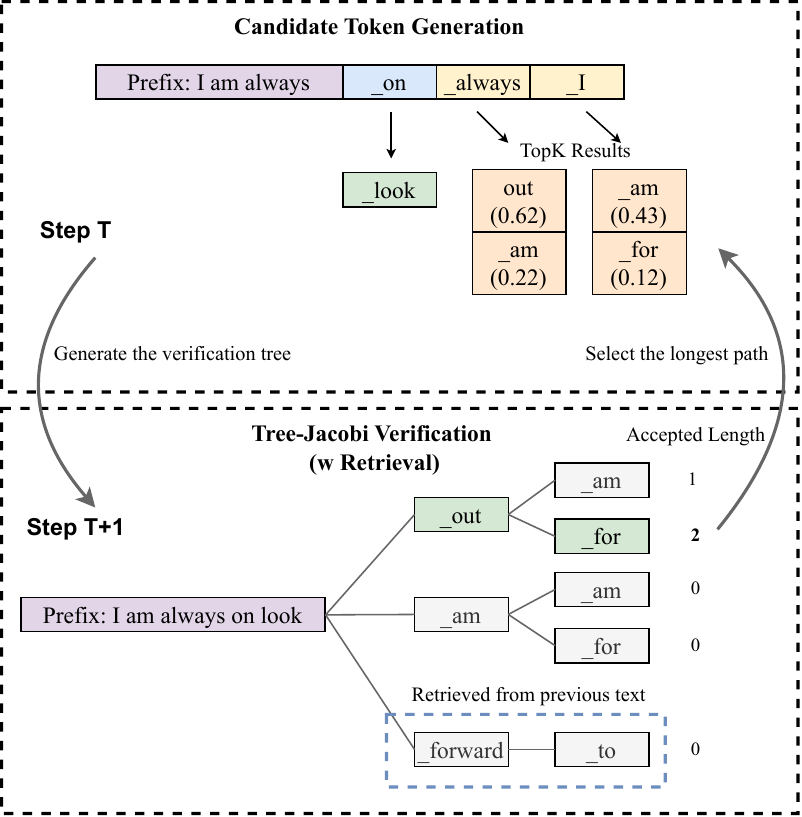}
  \caption {The main flowchart of TR-Jacobi decoding. It should be noted that candidate generation and tree verification are performed \textbf{in the same step}.
  For clarity, we choose candidate generation at moment T and tree verification at moment T+1 for analysis in the figure.}
  \label{fig:tree-based}
\end{figure}

\paragraph{Tree-based Jacobi Decoding.}
As discussed in Section \ref{tree-verify}, using token tree verification has become a common method of verification in speculative decoding.
In this paper, we also want to improve the efficiency of Jacobi decoding by constructing multiple candidate sequences.
Like Medusa \cite{cai2024medusa}, we heuristically chose a sparse tree as our tree-attention template (see the Appendix \ref{app:tree} for details).
At the beginning of the generation, we initialise all the nodes of the tree using ahead noise to start the tree-based Jacobi decoding.
As shown in Figure \ref{fig:tree-based}, for each forward process, each path performs an ordinary Jacobi decoding process via tree attention.
We then choose the longest accept-length path and continue to fill the validation tree nodes for next round based on the path's subsequent predictions.
It is important to note that we use the ahead noise tokens to populate the remaining positions in the validation tree, just like regular Jacobi decoding.

\paragraph{Retrieval-Augmented Jacob Decoding.}
In addition, for methods that design draft token predictions on the input side of the model (e.g., Jacobi, BiTA, etc.), cold-start is also a key issue that needs to be addressed.
When all draft tokens of this input are accepted, the model will have no way to get new draft tokens in this round.
Existing methods mitigate this problem by subsequently splicing more tokens, but incur additional inference costs.
To avoid starting validation from completely random noise in this case, we consider combining retrieval-based draft token and model-based draft token generation.

Specifically, we set a retrieval path in the token tree to hold the candidate tokens obtained by retrieving the previous tokens.
For retrieval, we use a simple and efficient method called prompt lookahead decoding \cite{saxena2023prompt} to obtain draft tokens with the same beginning directly from the current ahead tokens for verification, which significantly accelerates inference on tasks such as summarization.
The analysed experiments in Section \ref{pld} demonstrate that incorporating retrieved information is effective in improving the model's acceleration ratio in specific domains. Also, Jacobi decoding can alleviate the inherent problems of retrieval methods in domains such as translation.

\section{Experiments}

\subsection{Experimental Setup}
\label{sec:experiment}
\paragraph{Datasets.}
To verify that our proposed Make Some Noise (MSN) SFT training can bring inference acceleration without compromising model performance, we have constructed SFT datasets in the general and code domains, respectively.
For the general domain, we follow \citeposs{lin2024bita} setup to construct a training dataset containing 190k samples from LIMA \cite{zhou2024lima}, Alpaca-GPT4 \cite{peng2023instruction}, CodeAlpaca \cite{chaudhary2023code}, OpenPlatypus \cite{lee2023platypus} and CIP \cite{cip2023}.
Note that we only use 100k samples from CIP.
For the code domain, we adopt a total of 185k samples from Magicoder-OSS \cite{wei2023magicoder} and Evol-CodeAlpaca \cite{luo2023wizardcoder} as the training dataset, which are widely used in the program synthesis task.

\paragraph{Training Settings.}
To evaluate the proposed method comprehensively, we select LLama3-8B-Base \cite{touvron2023llama} and DeepseekCoder-6.7b-Base \cite{guo2024deepseek} as the foundation models for the general and code domains, respectively. The training settings for MSN are aligned with the baseline (SFT), maintaining a sequence length of 2048 tokens, a batch size of 512, and a training epoch of 4. Full-parameter fine-tuning is performed on two servers, each equipped with 8 A100-80GB GPUs, utilizing bf16 precision. We determine that a noise segment length of 4 is optimal for dynamic noise replacement for each sample.

\begin{table}[t]
    \centering
    \scalebox{0.9}{
    \begin{tabular}{lccl}
    \toprule
     \multirow{2}{*}{\textbf{Method}} &  \multirow{2}{*}{\textbf{Metric(+)}} & \textbf{Speed} & \multirow{2}{*}{\textbf{Speedup}} \\
      &   & \textbf{(tokens/s)} &  \\
    \midrule
    \multicolumn{4}{c}{LLaMA3-8B-Base (General)} \\
    \midrule
    SFT    & \multirow{3}{*}{\textbf{6.13}/60.38} & 38.00 & 1.00$\times$\\
    +Jacobi    & & 39.38 & 1.01$\times$\\
    +TR-Jacobi   & & 72.04 & 1.90$\times$\\
    \hdashline
    \textbf{MSN (Ours)} & \multirow{3}{*}{6.12/\textbf{60.87}} & 44.69 & 1.00$\times$ \\
    +Jacobi    & & 72.53 & 1.62$\times_{\uparrow60}$ \\
    +TR-Jacobi    & & 99.32 & 2.28$\times_{\uparrow20}$ \\
    \midrule
    \multicolumn{4}{c}{DeepseekCoder-6.7B-Base (Code)} \\
    \midrule
    SFT & \multirow{3}{*}{76.6 (68.6)} & 48.47 & 1.00$\times$ \\
    +Jacobi & & 48.59 & 1.00$\times$ \\
    +TR-Jacobi &  & 90.81 & 2.08$\times$ \\
    \hdashline
    \textbf{MSN (Ours)} & \multirow{3}{*}{\textbf{77.0} (\textbf{68.7})} & 48.03 & 1.00$\times$\\
    +Jacobi &  & 89.73 & 1.97$\times_{\uparrow97}$\\
    +TR-Jacobi &  & 128.50 & 2.68$\times_{\uparrow29}$\\
    \bottomrule 
    \end{tabular}
    }
    \caption{Results of task performance experiments in general and code domains.
    The general domain metric uses scores from MT-bench and MMLU-weighted from MMLU.
    The code domain uses pass@1 under greedy decoding.
    For the code domain, we choose the average of HumanEval and MBPP as a composite metric.
    `(+)': Results after executing additional tests from evalplus.
    `$\uparrow$': Percentage improvement over models without MSN.
}
    \label{tab:main_task}
\end{table}


\begin{table*}[t]
    \centering
    \scalebox{0.90}{
    \begin{tabular}{lcccccccccl}
    \toprule
    \multirow{2}{*}{\textbf{Method}} & \multirow{2}{*}{\textbf{AS}} & \multirow{2}{*}{\textbf{MT-B}} & \multirow{2}{*}{\textbf{Trans}} & \multirow{2}{*}{\textbf{Sum}} & \multirow{2}{*}{\textbf{QA}} & \multirow{2}{*}{\textbf{Math}} & \multirow{2}{*}{\textbf{RAG}} & \multirow{2}{*}{\textbf{\#MAT}} & \textbf{\#Speed} & \multirow{2}{*}{\textbf{Overall}} \\
     & & & & & & & & & \textbf{(tokens/s)} &  \\
    \midrule
    \multicolumn{11}{c}{Vicuna-7B-v1.3}\\
    \midrule
    AR & \ding{55} & 1.00$\times$ & 1.00$\times$ & 1.00$\times$ & 1.00$\times$ & 1.00$\times$ & 1.00$\times$ & 1.00 & 49.64 & 1.00$\times$\\
    PLD & \ding{55} & 1.60$\times$ & 1.03$\times$ & 2.58$\times$ & 1.15$\times$ & 1.72$\times$ & 2.15$\times$ & 1.85 & 84.23 & 1.69$\times$\\
    Medusa2 & 1.6B & 2.54$\times$ & 2.01$\times$ & 2.22$\times$ & 2.00$\times$ & 2.59$\times$ & 2.09$\times$ & 3.12 & 111.49 & \textbf{2.25}$\times$  \\
    EAGLE & 0.3B & 2.59$\times$ & 1.91$\times$ & 2.25$\times$ & 2.07$\times$ & 2.61$\times$ & 2.01$\times$ & 3.58 & 111.58 & \textbf{2.25}$\times$\\
    \hdashline
    LookAhead & \ding{55} & 1.44$\times$ & 1.14$\times$ & 1.31$\times$ & 1.26$\times$ & 1.57$\times$ & 1.21$\times$ & 1.65 & 65.80 & 1.32$\times$\\
    Jacobi & \ding{55} & 0.95$\times$ & 0.92$\times$ & 0.94$\times$ & 0.94$\times$ & 0.98$\times$ & 0.94$\times$ & 1.05 & 47.06 &  0.95$\times$\\
    TR-Jacobi & \ding{55} & 1.69$\times$ & 1.31$\times$ & 2.10$\times$ & 1.28$\times$ & 1.74$\times$ & 1.58$\times$ & 2.00 & 80.30 &  1.62$\times$\\
    \midrule
    \multicolumn{11}{c}{LLaMA3-8b-MSN \textbf{(Ours)}}\\
    \midrule
    AR & \ding{55} & 1.00$\times$ & 1.00$\times$ & 1.00$\times$ & 1.00$\times$ & 1.00$\times$ & 1.00$\times$ & 1.00 & 42.13 &  1.00$\times$ \\
    \hdashline
    LookAhead & \ding{55} & 1.51$\times$ & 1.36$\times$ & 1.46$\times$ & 1.35$\times$ & 1.65$\times$ & 1.40$\times$ & 1.75 & 61.51 & 1.46$\times_{\uparrow11}$ \\
    Jacobi & \ding{55} & 1.62$\times$ & 1.54$\times$ & 1.75$\times$ & 1.41$\times$ & 1.67$\times$ & 1.48$\times$ & 1.86 & 66.68 &  1.58$\times_{\uparrow66}$\\
    TR-Jacobi & \ding{55} & 2.22$\times$ & 2.03$\times$ & 2.77$\times$ & 1.85$\times$ & 2.16$\times$ & 1.96$\times$ & 2.94 & 91.63 & \textbf{2.17}$\times_{\uparrow34}$\\
    \bottomrule
    \end{tabular}
    }
    \caption{Experimental results of acceleration ratios in various areas of Spec-Bench (Multi-turn Conversation, Translation, Summarization, Question Answering, Mathematical Reasoning, Retrieval-aug. Generation).
Under the dashed line indicates the Jacobi-like decoding method.
`AS': Additional Structure.
`\#MAT': \#Mean Accepted Tokens.
`$\uparrow$': Percentage improvement over models without MSN.}
    \label{tab:acc_task}
\end{table*}

\paragraph{Evaluation Settings.}
In this paper, we conduct experiments on the task performance and acceleration performance of MSN, respectively.
For task performance, we use the MT-bench \cite{zheng2024judging} and MMLU \cite{hendrycks2020measuring} in the general domain, the HumanEval \cite{chen2021evaluating} and MBPP \cite{austin2021program} benchmarks in the code domain for evaluation. Evalplus \cite{evalplus} , which provides additional test cases for problems in HumanEval and MBPP, is also included. 
For acceleration performance, we performed a speedup evaluation of the proposed parallel decoding methods on Spec-Bench \cite{xia2024unlocking}. This benchmark contains data from multiple domains and provides a fair comparison with existing acceleration methods.
Following previous work, all speed related experiments are done on a single A100-80G device with the batch size as 1.
For our MSN model, the draft token length during inference is consistent with the noise segment length during training, which is 4.

\subsection{Comparison with SFT}
\paragraph{Baselines.}
We first validate the impact of the proposed MSN training framework on the performance of the model tasks in the general and code domains.
The standard supervised fine-tuning (SFT) is chosen as the baseline method for comparison.
Specifically, we perform domain-specific SFT and noise training based on the same base model and compare the performance of both on downstream tasks.

\paragraph{Results.}
The metric in Table~\ref{tab:main_task} represents the task performance of each model.
There is no significant performance loss of the model trained by MSN on the downstream task compared to SFT.
Futhermore, The MSN model even delivers a slight performance boost in both domain. 
The enhancement in the code domain is particularly noteworthy, given that evaluating generated programs is more rigorous than evaluating conversation.
Programs must be correctly formatted and pass all test cases to be deemed successful. It indicates that MSN does not hurt the model to acquire capabilities during the SFT phase.
Our analysis suggests that this gain comes from the fact that noise mitigates the negative effects of teacher forcing training on the model to some extent.
The causal denoising task forces the model to focus on more distant tokens when predicting the current location token because the current input is noisy.
We also conduct som experiments comparing SFT and MSN with different base models in the code domain, which can be found in Appendix \ref{app:code}.

In addition to this, we briefly test the acceleration effect of the MSN method on the Jacobi-like decoding strategy.
We can see that targeted training on the denoising ability of the model significantly improves the acceleration ratio of Jacobi decoding in different domains.
Our proposed TR-Jacobi further improves the acceleration ratio by verifying multiple paths simultaneously.

\subsection{Comparison with Other Speculative Decoding Methods}
\paragraph{Baselines.}
To further compare MSN with existing speculative decoding methods, we conducted an evaluation on Spec-Bench \cite{xia2024unlocking}.
We choose both speculative methods that include no additional structures (Jacobi, LookAhead, PLD) and those that require additional structures (Medusa2, EAGLE) for comparison.
EAGLE and Medusa2 are post-trained on Vicuna-7b-v1.3 \cite{vicuna2023}, which is already a post-SFT model. Since our proposed MSN is performed in the SFT stage, we need to perform MSN SFT on a base model.
Therefore, we conduct MSN on LLaMA3-8B-Base and perform acceleration evaluations on two different foundation models for a rough comparison of speedup ratios based on different auto-regressive (AR) throughputs.


\paragraph{Results.}
The overall acceleration experiment results are shown in Table \ref{tab:acc_task}.
After specific training on denoising capabilities, the MSN model improves the speedup ratio on all Jacobi-like decoding strategies.
For LookAhead, the denoising ability may produce incoherent n-grams, which can lead to a relatively low improvement.
For both Jacobi decoding and TR-Jacobi decoding acceleration ratios, noisy training brings significant improvements.
TR-Jacobi has a fine blend of retrieved and generated draft tokens with respectable average receive lengths in all domains.

 The speedup ratio of the MSN model under TR-Jacobi decoding is competitive with other methods.
As a method with no additional training stages and no additional model structure, the proposed acceleration method is also comparable to the models with additional structures.
It is fair to say that MSN is a lightweight and efficient way to achieve inference speedup comparable to existing SOTA models while improving model robustness.


\section{Discussion}

\begin{table}[t]
    \centering
    \scalebox{0.9}{
    \begin{tabular}{ccccc}
    \toprule
    \multirow{2}{*}{L}  & \multirow{2}{*}{HEval(+)} & \multirow{2}{*}{MBPP(+)} & Speed & \multirow{2}{*}{Speedup} \\
      & & & (tokens/s) &  \\
    \midrule
    1 & \textbf{74.4} (\textbf{70.1}) & 75.9 (64.6) & 58.35 & 1.55$\times$ \\
    4 & 73.2 (68.3) & \textbf{76.5} (64.3) & 80.47 & \textbf{2.13}$\times$ \\
    8 & 71.3 (65.9) & 76.5 (\textbf{65.1}) & 80.02 & 2.12$\times$ \\
    \bottomrule
    \end{tabular}
    }
    \caption{The effect of the training noise segments length on acceleration and task capability.
    `L' represents the length of the noise segment.}
    \label{tab:train_span_leng}
\end{table}

\begin{figure}[t]
    \centering
    \centering
    \includegraphics[width=0.8\linewidth]{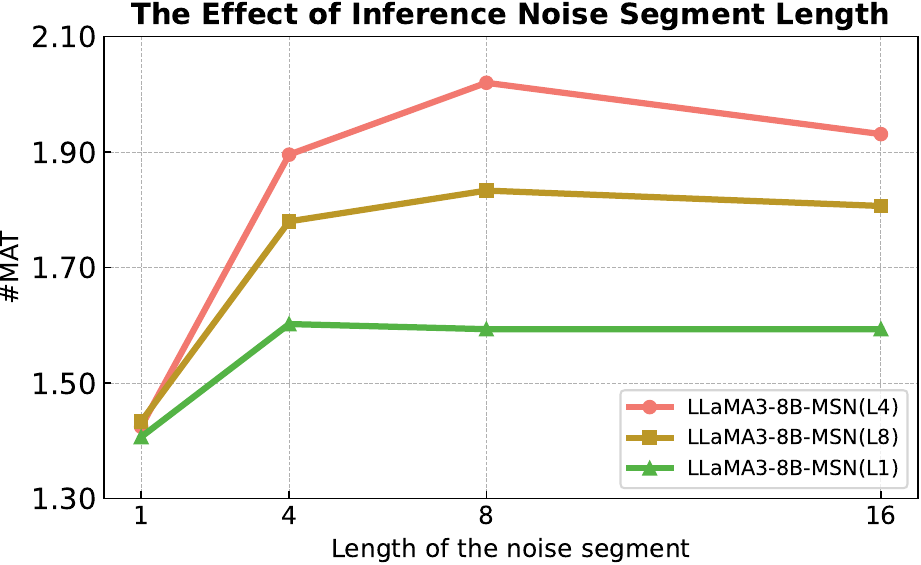}
    \caption{The effect of the inference noise segments length on acceleration with Jacobi decoding.
    `\#MAT': Mean Accepted Token.}
    \label{fig:inference_span_len}
\end{figure}

\begin{figure*}[t]
    \centering
    \includegraphics[width=\textwidth]{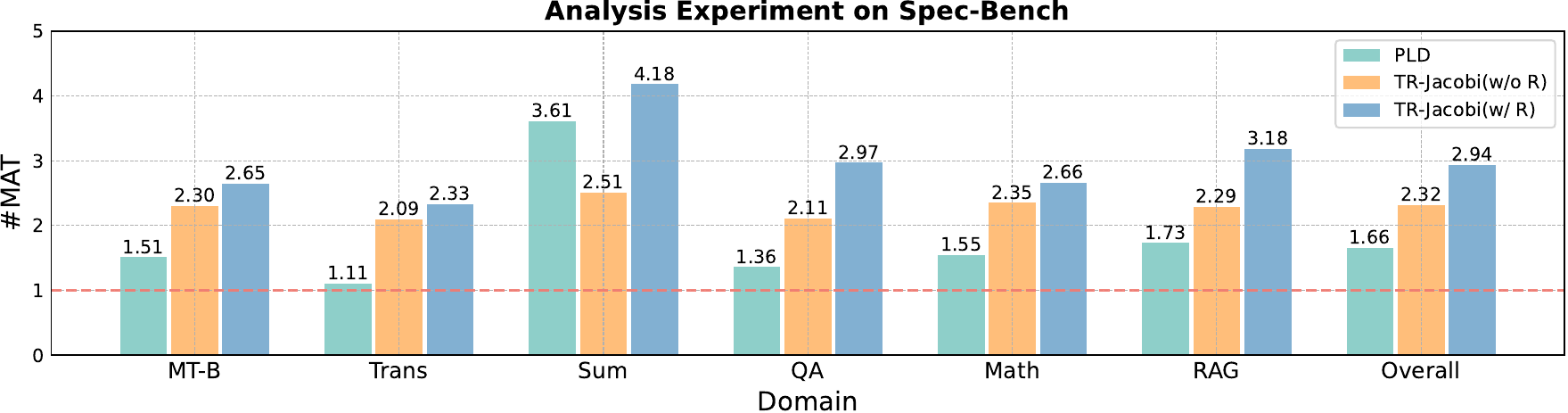}
    \caption{Results of ablation experiments on the retrieval part of TR-Jacobi decoding.}
    \label{fig:retrieval}
\end{figure*}

\begin{figure}[t]
    \centering
    \includegraphics[width=0.8\linewidth]{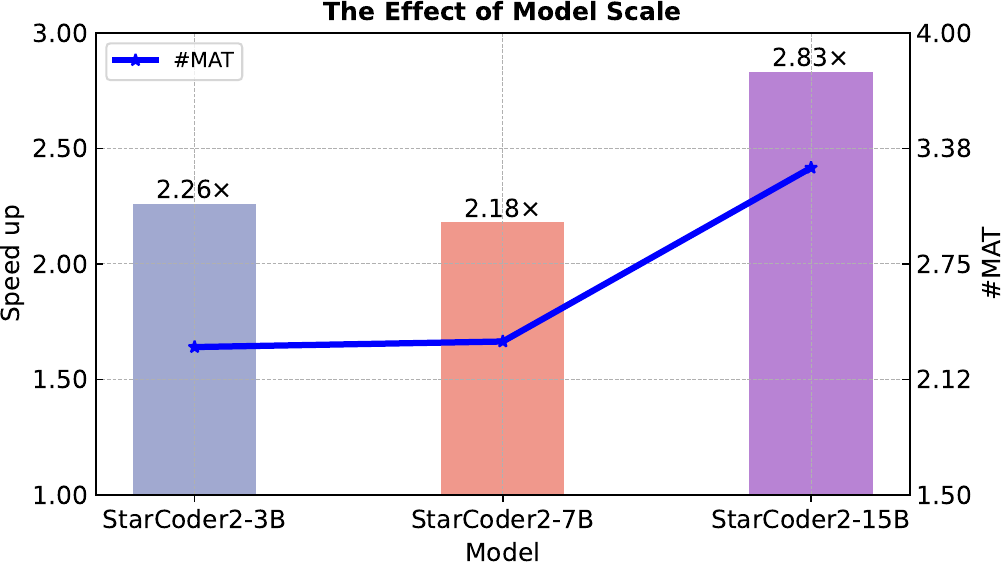}
    \caption{Acceleration experimental results of MSN training for StarCoder2 models of different sizes.}
    \label{fig:model-scale}
\end{figure}

\subsection{Effect of Noise Segment Length}
\label{span-len}
The span length includes the length of the noise segment during training and the length of the draft sequence added during inference. The training span length affects the difficulty of the model learning from samples, while the span length during inference impacts both the hit length and the speculative operation latency.
\paragraph{Training Noise Segment length.}
\label{train_span}
Traning Noise segment length refers to the number of noise tokens. If the length is too short, the denoising capability of the model may be diminished, resulting in limited acceleration during inference. Conversely, if the length is too long, it significantly increases the difficulty of denoising, affecting the model's understanding of the sample and thereby harming its task performance. To observe the impact of varying training span lengths, we experiment with span lengths of 1, 4, and 8 on Deepseek Coder and the task performance and acceleration are shown in Table \ref{tab:train_span_leng}. It demonstrates that a length of 1 yields high task performance but offers minimal acceleration. A length of 8 provides substantial acceleration but at the cost of significant task performance degradation. A length of 4 achieves the highest acceleration with a lower impact on performance.

\paragraph{Inference Noise Segment Length} 
Inference noise segment length represents the draft token num for Jacobi iteration, which is also the maximum number of times the token can be iteratively denoised.
We perform parallel inference experiments with different inference noise segment lengths for models trained with different training noise segment lengths (described above).
We find that the model can generalize from a smaller training noise segment size to a larger inference noise segment size.
This suggests that even though we only trained one step to go directly from noise token to gold token, the model is able to generalize to obtain iterative denoising ability.
In addition, the training noise length of 8 does not outperform the training noise length of 4, suggesting that length 4 has reached the bottleneck of the model's denoising ability in the SFT stage.

\subsection{Effect of Model Scale}
To assess the generalisation capability of MSN, experiments are conducted on different sizes of Starcoder2 \cite{lozhkov2024starcoder}, specifically 3B, 7B, and 15B parameters.
The training data remains consistent with Section \ref{sec:experiment}, and HumanEval with Jacobi decoding is utilised to evaluate the acceleration. 
The results of the experiment are shown in Figure \ref{fig:model-scale}.
Overall, MSN demonstrates significant speedup across all model sizes, indicating its broad applicability.

Specifically, when increasing the model size from 3B to 7B, the Mean Accepted Tokens (\#MAT) only increases by 0.03, and the speedup ratio slightly decreases. It suggests that a 3B model is sufficient to learn the denoising capability and that the effectiveness of denoising does not significantly change with an increase in parameters from 3B to 7B. The incremental increase in MAT for the 7B model is insufficient to offset the additional computational cost of draft tokens during inference, resulting in a decrease in the speedup ratio. However, when the model size reaches 15B, the denoising capability increases dramatically. The \#MAT rises by nearly 1, and the additional computational cost of draft tokens is mitigated by the substantial improvement in hit rate, resulting in a 0.6 increase in the speedup ratio. The outcomes on model scale further exemplify the extensive applicability of our method and demonstrate that larger models have greater potential.


\subsection{Effect of Retrieval Paths}
\label{pld}
In order to further analyse the performance enhancement brought by the retrieval paths to TR-Jacobi decoding, we perform ablation experiments with Llama3 on Mt-Bench.
We compare the \#MAE for the pure retrieval method PLD, the pure Jacobi method TR-Jacobi w/o R, and TR-Jacobi on each domain.
The results of the experiment are shown in Figure \ref{fig:retrieval}.
Our proposed TR-Jacobi integrates and surpasses pure Jacobi and pure retrieval solutions in terms of acceleration performance in various domains.
Retrieval paths mitigate the cold start and instability due to random noise of Jacobi's approach.
The Jacobi method can continue to iterate over the retrieval path and can also handle tasks with shorter contexts (e.g., translation).

\section{Conclusion}
In this paper, we propose an effective training framework Make Some Noise (MSN) to be used as a replacement for the SFT stage.
It enhances the denoising ability of the model without affecting the SFT training performance.
Combined with our proposed TR-Jacobi decoding strategy, the MSN model is able to achieve 2.3-2.7x speedup in the general and code domains without additional structure and training.


\section*{Limitations}
Causal denoising, as a more general task, is only used for experiments in the SFT phase in this paper due to limited computational resources.
It is a worthy exploration to merge the denoising task with the next token prediction task into the pre-training task.
In addition to this, the optimal noise fragment length may be related to the content of the SFT training set (parallel prediction of code text is less difficult, natural language text is more difficult).
For a new SFT dataset, confirming the optimal noise segments may require some pre-experiments for searching, which imposes a certain burden on MSN training.


\section*{Ethics Statement}
The source data for proposed methods come exclusively from publicly available project resources on legitimate websites and do not involve any sensitive information.
In addition, all baselines and datasets used in our experiments are also publicly available, and we have acknowledged the corresponding authors by citing their work.

\section*{Acknowledgements}
We gratefully acknowledge the support of the National Natural Science Foundation of China (NSFC) via grant 62236004, 62206078, 62441603 and 62476073 and the support of Du Xiaoman (Beijing) Science Technology Co., Ltd.

\bibliography{acl_latex}

\appendix

\clearpage

\section{PPL-Based Location Selection}
\label{app:location}
As discussed in Section \ref{noisy_training}, we try to use the PPL-based selection method to select the location of the noise segments.
Inspired by \citet{lin2024rho}, different tokens contribute differently to the learning of that sample.
Therefore, we consider using cross-entropy loss to score the input tokens and select the segment with the lowest loss for noise replacement, which can be formulated as:
\begin{equation}
  \label{eq:location_noise}
  k = \arg\min{\sum_{i=k}^{k+l}-\log P(X_i|X_{<i};\theta)}
\end{equation}
where $k$ represents the start index of the noise segment and $l$ represents the length of the noise segment.
Segments with low cross-entropy loss possess both correct prediction and high prediction confidence.
Correct predictions indicate that the model has learnt this segment sufficiently and replacement with noise has minimal impact on model performance.
High prediction confidence means that the segment is likely to be a commonly used expression \cite{sun2024decoding}, which is useful for learning acceleration capabilities.

The final results of the experiment are shown in Table \ref{tab:seg_loc}. Even in code domains with stringent output requirements, ppl-based position selection has no significant speed or performance advantage over random selection.
Considering that the ppl-based training method is too complicated and increases the training time to some extent, we subsequently adopt random noise locations.

\begin{table*}[b]
    \centering
    \begin{tabular}{lcccccc}
    \toprule
         & \textbf{HumanEval (+)} & \textbf{MBPP (+)} & \textbf{Speed (tokens/s)} & \textbf{Speedup} \\
    \midrule
    Baseline & 77.4 (72.6) & 75.7 (64.6) & 44.01 & 1.00$\times$ \\
    Random & 76.8 (72.0) & 75.4 (65.1) & 99.96 & 2.11$\times$ \\
    PPL-Based & 77.4 (70.7) & 76.5 (66.7) & 101.18 & 2.13$\times$ \\
    \bottomrule
    \end{tabular}
    \caption{The comparison between the randomly selected noise segment and the lowest loss noise segment.}
    \label{tab:seg_loc}
\end{table*}

\section{Templates for Token Tree}
\begin{figure}[t]
  \centering
  \includegraphics[width=\linewidth]{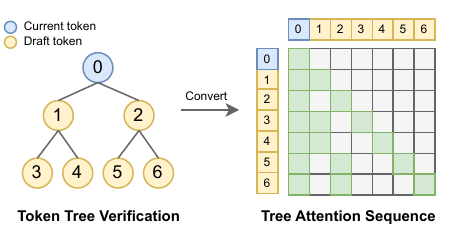}
  \caption {Illustration of token tree verification.
  The model achieves simultaneous verification of multiple candidate paths through a specially constructed sparse attention matrix.}
  \label{fig:tree-attention}
\end{figure}

As shown in Figure \ref{fig:tree-attention}, token tree verification organizes multiple paths into a tree structure, which is verified in parallel by sparse attention masks.
With high accuracy of draft token prediction, token tree verification can effectively improve the average acceptance length.
However, for Jacobi decoding, since no additional structure is introduced, the correct prediction rate of its draft token is relatively low, and the generation of draft fragments is mainly achieved by iterative decoding.
Therefore the enhancement brought by tree verification mainly depends on the topK of the first draft token, and experiments show that TR-Jacobi decoding is not sensitive to the structure of the verification tree.
In this paper, we use the same heuristic tree structure as vicuna-7b in medusa \cite{cai2024medusa}, containing 63 nodes.
In particular, we also add a retrieval path of length 5 to store the retrieved draft tokens.

\label{app:tree}

\section{More Experiment Results}
\label{app:code}
In addition to the previously mentioned DeepSeek-Coder, we have experimented on a variety of different code base models (CodeLlama \cite{roziere2023code}, StarCoder \cite{li2023starcoder}).
The results of the experiments on Humaneval and MBPP are shown in Table \ref{tab:code_mode}.
We can see that the MSN and SFT methods do not reflect a significant gap on models of different sizes and sources.


\begin{table*}[t]
    \centering
    \begin{tabular}{lccccc}
    \toprule
    \textbf{BaseModel} & \textbf{Method} & \textbf{Humaneval} & \textbf{Humaneval+} & \textbf{MBPP} & \textbf{MBPP+}\\
    \midrule
    CodeLlama 7B & SFT & 62.20 & \textbf{57.90} & \textbf{65.60} & \textbf{57.70}\\
    CodeLlama 7B & MSN & \textbf{64.00} & \textbf{57.90} & 63.50 & 54.50\\
    StarCoder 3B & SFT & 57.30 & 53.70 & \textbf{60.60} & 52.40\\
    StarCoder 3B & MSN & \textbf{59.10} & \textbf{55.50} & 60.30 & \textbf{52.60}\\
    StarCoder 7B & SFT & \textbf{68.90} & \textbf{64.60} & 63.80 & 55.00\\
    StarCoder 7B & MSN & 66.50 & 60.40 & \textbf{64.30} & \textbf{55.60}\\
    \bottomrule
    \end{tabular}
    \caption{Experimental results for different code base models.}
    \label{tab:code_mode}
\end{table*}

\begin{table*}[t]
    \centering
    \begin{tabular}{cccccccc}
    \toprule
     \textbf{Method} & \textbf{humanities} & \textbf{stem} & \textbf{social-science} & \textbf{other} & \textbf{MMLU} & \textbf{MMLU-weighted}\\
    \midrule
    SFT & 65.18 & \textbf{51.74} & 69.72 & \textbf{64.71} & 61.55 & 60.38 \\
    MSN & \textbf{66.32} & 51.20 & \textbf{71.09} & 64.33 & \textbf{61.83} & \textbf{60.87} \\
    \bottomrule
    \end{tabular}
    \caption{Complete experiment results on MMLU Benchmark. Both SFT and MSN methods are trained on Llama3-8B-Base model.}
    \label{tab:mmlu}
\end{table*}


    

\end{document}